\documentclass[runningheads]{llncs}
\usepackage[T1]{fontenc}
\usepackage{amsmath}
\usepackage{booktabs}
\usepackage{cite}
\usepackage{xcolor}

\usepackage{pifont}% http://ctan.org/pkg/pifont
\newcommand{\cmark}{\ding{51}}
\newcommand{\xmark}{\ding{55}}

\usepackage{graphicx,verbatim}
\begin{document}

% NOTES on the manuscript:
% > Manuscripts can be up to 8 pages (text, figures, tables, conclusion and acknowledgement sections) plus up to 2 pages of references. 

\title{NICO-RAG: Multimodal Hypergraph Retrieval-Augmented Generation for Understanding the Nicotine Public Health Crisis} 
% \begin{comment}  %% Removed for anonymized MICCAI submission
\author{
    Manuel Serna-Aguilera\inst{1} \and
    Raegan Anderes\inst{1} \and
    Page Daniel Dobbs\inst{2} \and
    Khoa Luu\inst{1}
}
\institute{
    University of Arkansas, Fayetteville, AR, 72701, USA\\
    \email{\{mserna, rmandere, khoaluu\}@uark.edu} \and 
    University of Arkansas for Medical Sciences, Little Rock, AR, 72205, USA\\
    \email{PDDobbs@uams.edu}
}

% \end{comment}

% \author{Anonymized Authors}  %% Added for anonymized MICCAI submission
% \authorrunning{Anonymized Author et al.}
% \institute{Anonymized Affiliations \\
%     \email{email@anonymized.com}}
  
\maketitle              % typeset the header of the contribution
\begin{abstract}

The nicotine addiction public health crisis continues to be pervasive. In this century alone, the tobacco industry has released and marketed new products in an aggressive effort to lure new and young customers for life. Such innovations and product development, namely flavored nicotine or tobacco such as nicotine pouches, have undone years of anti-tobacco campaign work. Past work is limited both in scope and in its ability to connect large-scale data points. Thus, we introduce the \textbf{N}icotine \textbf{I}nnovation \textbf{C}ounter-\textbf{O}ffensive (\textbf{NICO}) Dataset to provide public health researchers with over 200,000 multimodal samples, including images and text descriptions, on 55 tobacco and nicotine product brands. 
In addition, to provide public health researchers with factual connections across a large-scale dataset, we propose \textbf{NICO-RAG}, a retrieval-augmented generation (RAG) framework that can retrieve image features without incurring the high-cost of language models, as well as the added cost of processing image tokens with large-scale datasets such as NICO. At construction time, NICO-RAG organizes image- and text-extracted entities and relations into hypergraphs to produce as factual responses as possible. This joint multimodal knowledge representation enables NICO-RAG to retrieve images for query answering not only by visual similarity but also by the semantic similarity of image descriptions. Experimentals show that without needing to process additional tokens from images for over 100 questions, NICO-RAG performs comparably to the state-of-the-art RAG method adapted for images. 
%In addition, to provide public health researchers with factual connections across a large-scale dataset, we propose \textbf{NICO-RAG}, a retrieval-augmented generation (RAG) that organizes both text descriptions and images of individual products within NICO. \textcolor{red}{[Add 1 sentence, to concisely and precisely summarize the novelty in NICO-RAG, in terms of the novelty of methodology of the proposed method, not just application.]} More precisely, NICO-RAG goes beyond only relying on image embeddings for retrieval, but instead performs knowledge retrieval using several descriptor functions, each giving useful information on products encoded in natural language, such as color, shape, and text present on packaging, all of which Large Language Models can exploit, to great effect. It allows NICO-RAG to retrieve images for query answering not only by visual similarity but also by visual semantic similarity. Our method aims to answer questions in a factual and informed manner, thereby vastly reducing the time required to manually make connections. With this, we hope to speed up research and policy efforts in public health. \textcolor{red}{[Add one sentence, to summarize the experimental results]}

%\textcolor{red}{Code is provided.}

% Authors must provide keywords and are not allowed to remove this Keyword section.

\end{abstract}

\section{Introduction} \label{sec:intro} % ~1.5 pgs

\begin{figure}[htp]
    \centering
    % Define the box dimensions: e.g., width=\textwidth, height=5cm
    %\framebox[\textwidth]{\rule{0pt}{5cm}}
    \includegraphics[width=1.0\linewidth]{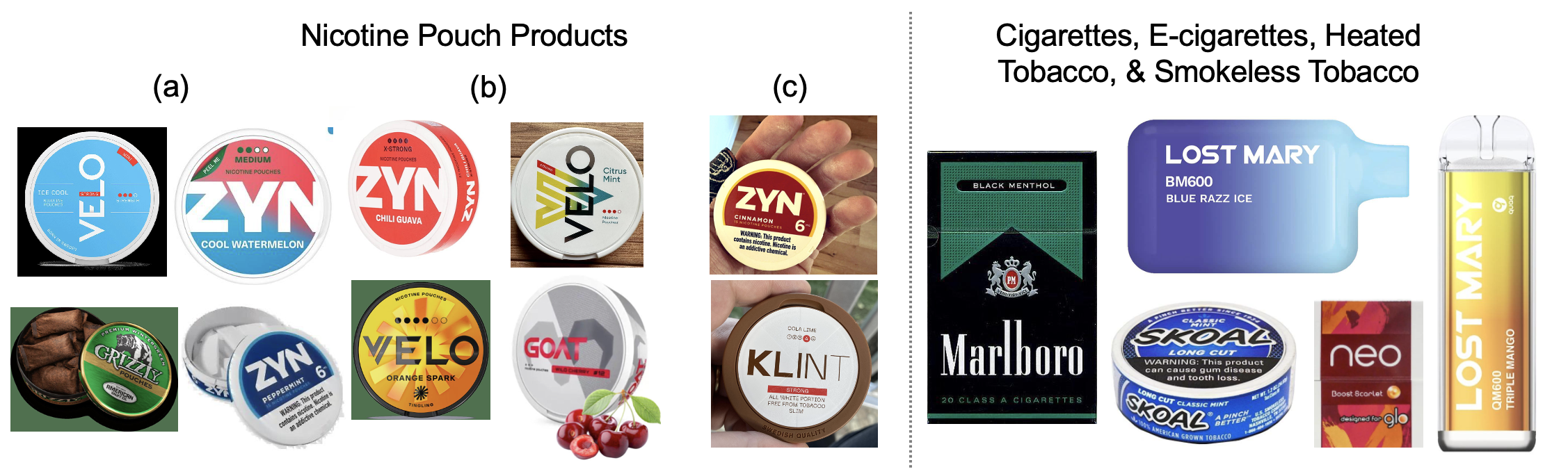}
    \caption{
        Samples of our NICO Dataset. (Left) Nicotine pouch samples with (a) mint-based flavors, (b) fruit flavors, and (c) spice and coffee flavors. (Right) Other tobacco and nicotine products with flavorings. 
        %\textcolor{orange}{[Manuel - I already thought of a better visualization. This will be replaced.]}
    }
    \label{fig:dataset-vis}
\end{figure}

% I) Introduction to the problem as we understand and as told by the public health side
In recent times, the tobacco industry has rapidly innovated in nicotine delivery and is releasing many new products on a massive scale. By leveraging legal loopholes, lobbying, court battles, slow legal proceedings, and a lack of awareness, tobacco companies can aggressively market new nicotine products, particularly towards youth, with little repercussions. Consequently, this rapid mass release of new, increasingly popular nicotine pouch products makes it very difficult for public health researchers and policy makers to address these innovations and marketing effectively. Traditionally, researchers have had to \textit{manually} sort, organize, and analyze large-scale datasets. This slow manual processing prolongs and exacerbates the nicotine addiction public health crisis. Therefore, there is a critical need to process large-scale data, e.g., images and documents, and to connect data points more quickly than current manual methods. 
% [Not needed—can remove to create space] Thus, we tackle the question: \textit{how can we process, organize, and understand massive amounts of data, and then provide users with meaningful results that draw connections between old data points and new data}? We aim to support public health research to make fine-grained and broad connections between both current and older products, with the end goal of closing the knowledge gap between tobacco company efforts and legislation. Without a fine-grained and broad understanding of how new products are used to lure young customers, a new generation of people will become addicted with this new class of nicotine packaging. 

% II) What are the existing approaches that are done? What is important to extract? What is our data domain (images, text, etc.)?
Current research that seeks to understand tobacco and nicotine products is rather limited in scope and capacity. Datasets from works such as Vassey et al. \cite{vassey2024scalable-tobacco} analyze 6,999 Instagram images labeled for e-cigarette-related objects for product detection using dynamic head attention \cite{dai2021dynamic}. Murthy et al. \cite{murthy-tobacco-dataset-2024} annotated 826 frames from TikTok videos and detected e-cigarette device use with a YOLOv7 detector \cite{wang2023yolov7}. In larger-scale work, Chappa et al. \cite{chappa2024phad} perform product classification on various types of products using video frames from TikTok and YouTube. Despite this progress, there remain gaps between what the AI side can deliver and what the public health side needs. To our knowledge, no dataset or methodology captures key relationships within large-scale data to enable informed, factual retrieval of vital product information, e.g., brand identifiers, distinctive visual features, flavors, advertising strategies, etc., for both the same product type and new and upcoming products.

% III) What do we propose? Summarize our contributions
\textbf{Contributions of this Work:}
We therefore address two large-scale problems, i.e., the need for a large dataset to build a unified knowledge base and, consequently, a methodology to leverage the connections between data points implicit in such a dataset. In this work, we make three main contributions. We first introduce the large-scale Nicotine Innovation Counter--Offensive (NICO) Dataset, comprising over 200,000 images of several product types across tobacco and nicotine product brands, with samples shown in Figure \ref{fig:dataset-vis}. Second, we propose the Nicotine Innovation Counter-Offensive Retrieval-Augmented Generation (NICO-RAG), a novel multi-modal and multi-feature framework for retrieval-augmented generation (RAG) in public health scenarios. We built NICO-RAG to handle combined text-and-image queries and not rely solely on image embeddings, but instead on an enriched combination of multiple descriptors of nicotine products—visual features, text embeddings, words on product packaging, color, morphology, etc. Finally, these simple product descriptors enable more holistic and diverse knowledge construction and retrieval processes while avoiding reliance on large, expensive language models for large-scale work. It enables public health researchers to more accurately link products across multiple criteria at scale in a fraction of the time.

\section{The Nicotine Innovation Counter-Offensive Dataset}

\begin{table*}[tb] 
\centering
\caption{
    A comparison of the past dataset with our proposed dataset. Our dataset contains 55 named tobacco or nicotine brands, 47 more than the PHAD \cite{chappa2024phad}. We also contribute not only images and labels but also textual, color, and shape descriptions. Finally, our dataset contains more diverse images than previous works.
}
\label{tab:dataset-comparison}
\begin{tabular}{lccccc}
\toprule
\textbf{Dataset} & \textbf{\begin{tabular}[c]{@{}c@{}}Product\\ Brands\end{tabular}} & \textbf{\begin{tabular}[c]{@{}c@{}}Text\\ Descriptions\end{tabular}} & \textbf{\begin{tabular}[c]{@{}c@{}}Color\\ Descriptions\end{tabular}} & \textbf{Shape} & \textbf{\begin{tabular}[c]{@{}c@{}}Number of\\ Image Samples\end{tabular}} \\
\midrule
Murthy et al. \cite{murthy-tobacco-dataset-2024} & 3 & \textcolor{red}{\xmark} & \textcolor{red}{\xmark} & \textcolor{red}{\xmark} & 826 \\
Vassey et al. \cite{vassey2024scalable-tobacco} & 7 & \textcolor{red}{\xmark} & \textcolor{red}{\xmark} & \textcolor{red}{\xmark} & 6,999 \\
PHAD \cite{chappa2024phad} & 8 & \textcolor{red}{\xmark} & \textcolor{red}{\xmark} & \textcolor{red}{\xmark} & 171,900 \\
\textbf{NICO (Ours)} & \textbf{55} & \textcolor{blue}{\cmark} & \textcolor{blue}{\cmark} & \textcolor{blue}{\cmark} & \textbf{202,599} \\
\bottomrule
\end{tabular}
\end{table*}

The Nicotine Innovation Counter-Offensive Dataset is the first of our two main contributions to advancing monitoring and understanding of tobacco and nicotine product innovation. With over 200,000 samples, it represents the largest and most diverse dataset of tobacco and nicotine products assembled. NICO comprises images and natural-language descriptors—coloring, text on packaging, simple color descriptions, product shape, and a simple description. Our dataset contributions are summarized in Table \ref{tab:dataset-comparison}. In contrast, prior work by Vassey et al. \cite{vassey2024scalable-tobacco} and Murthy et al. \cite{murthy-tobacco-dataset-2024} analyzes 6,999 and 826 e-cigarette images, respectively. The PHAD \cite{chappa2024phad} video dataset comprises 171,900 video frames across 8 brands, with no product descriptors and a smaller label set. 

% How did we collect our dataset; what was the **collection** process like?
\noindent
\textbf{Image Data Collection.} A significant portion of the images in NICO was collected through a rigorous process of sampling product images from various online resources and product catalogs. The image collection involved a combination of automated and manual processes. We first collected data using the Apify platform \cite{apify-citation}, which enabled query-based categorization. This categorization initially assigned labels for tobacco type, product type, and brand. In addition, we filtered out as many irrelevant samples as possible to ensure quality control, yielding a large-scale image set of products, sorted by tobacco type, product type, and brand hierarchical labels, which we use to organize every sample. 

% How do we prepare the dataset to use with our method? How do we process each sample—image, text, etc. How do we prepare the questions? What kinds of questions do we ask and why?
\noindent
\textbf{Data Preparation.} With the images collected, we prepared our image modules to provide natural-language information about the products (if any) in the images, without relying solely on Large Language Models (LLMs) or Large Multimodal Models (LMMs). We focus on the following image descriptors: natural language, optical character recognition (OCR), color, and shape. Further details for descriptor extraction formulation and implementations are provided in Sections \ref{sec:methodology} and \ref{subsec:experimental-design}. Additionally, in collaboration with our team of public health researchers, we hand-crafted questions on hot topics in nicotine pouch product research. There are 11 unique questions covering topics such as the flavors brands offer, the relationship between flavors and colors, the relationship between current pouch flavors and other product types, etc. 
Further details on experimental settings are given in Section \ref{subsec:experimental-design}.  % necessary??
% >>> There are XXX unique questions in NICO. The topics of these questions are X, Y, and Z.

\noindent
\textbf{NICO Dataset Statistics.} The NICO dataset contains a total of 202,599 image samples spanning 55 tobacco or nicotine products. The images span five product types that have undergone or are undergoing innovation, i.e., cigarettes, heated tobacco, e-cigarettes, smokeless tobacco, and nicotine pouches. Our dataset contains 50,882 cigarette images (13 brands), 48,261 heated tobacco images (14 brands), 61,243 e-cigarette images (11 brands), and 42,213 smokeless tobacco and nicotine pouch images (17 brands).

%%%%%%%%%%%%%%%%%%%%%%%%%%%%%%%%%%%%%%%%%%%%%%%%%%

%\section{NICO-RAG Methodology \textcolor{red}{[Full name of NICO-RAG]}} \label{sec:methodology} % ~3 pgs; we can compress this section by making some equations in-text instead of in the equation block.

\section{Nicotine Innovation Counter-Offensive Retrieval Augmented Generation (NICO-RAG) Approach} \label{sec:methodology}

\begin{figure}[tp]
    \centering
    % Define the box dimensions: e.g., width=\textwidth, height=5cm
    %\framebox[\textwidth]{\rule{0pt}{6cm}}
    \includegraphics[width=1.0\linewidth]{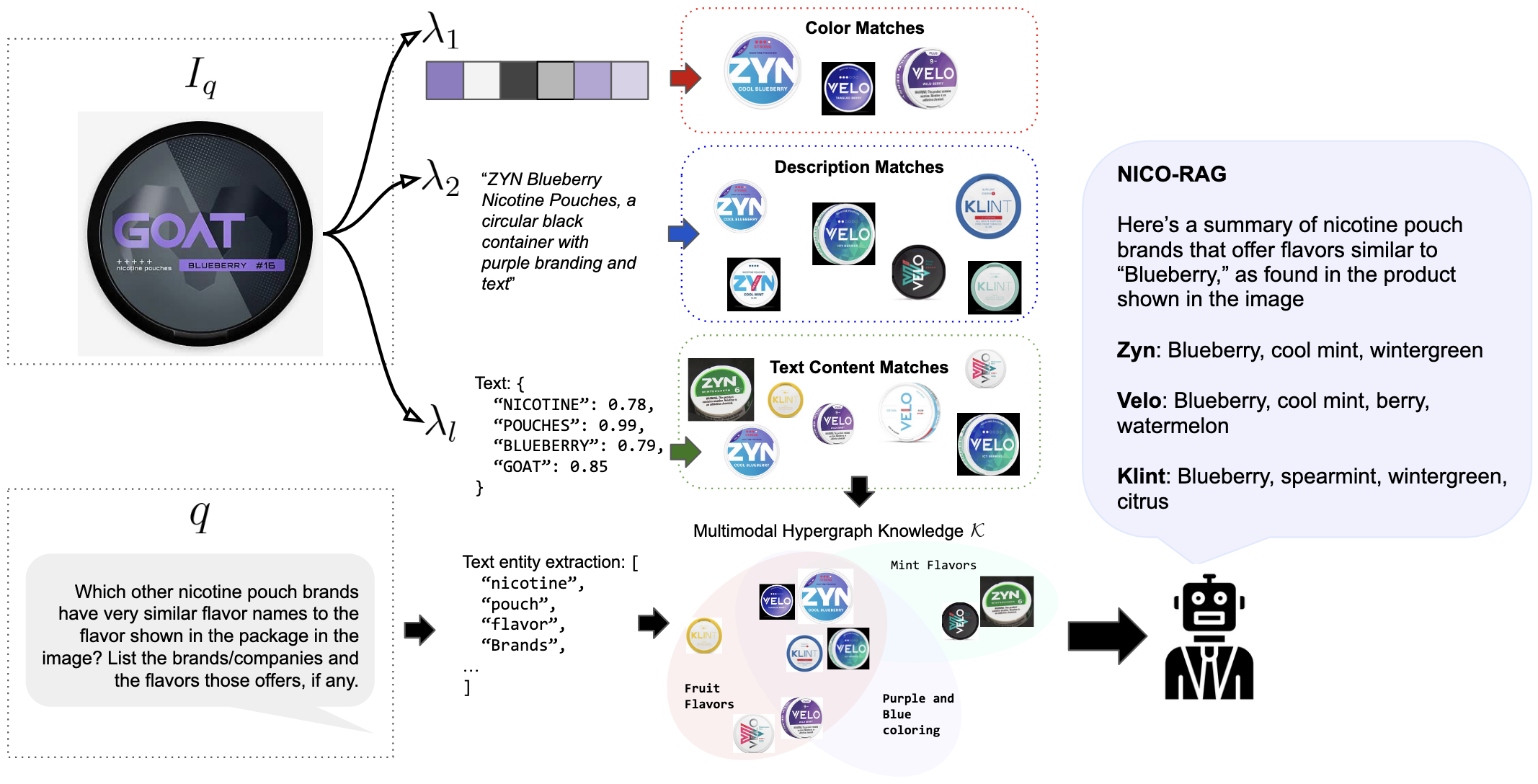}
    \caption{
        The NICO-RAG framework. We take in a query image and text, and via text and image entity discovery, we create the multimodal hypergraph knowledge $\mathcal{K}$, giving us image descriptors to capture all aspects desirable for public health in tobacco and nicotine prevention research. 
        %\textcolor{orange}{[Manuel - I already thought of a better visualization, but will take this weekend to make. This will be replaced.]}.
    }
    %Our proposed multimodal retrieval-augmented generation methodology, where we construct and retrieve knowledge based on image descriptors critical for tobacco or nicotine product analysis.}
    \label{fig:method-vis}
\end{figure}

\subsection{Preliminaries}

We first provide preliminary definitions for NICO-RAG, following those in previous work \cite{luo2025hypergraphrag}. The multimodal knowledge is denoted by $\mathcal{K}$ and is organized into text chunks ($K_{\text{chunk}}$)---a graph or hypergraph representation. In a graph or hypergraph setting, $\mathcal{K}$ is defined by entities and edges in Eqn. \eqref{eq:knowledge-def}.

\begin{equation} \label{eq:knowledge-def}
    \mathcal{K} = \Big(V^{(M)}, E_H^{(M)} \Big) = \Big(V^{(I)} \cup V^{(T)}, E^{(I)} \cup E^{(T)}\Big)
\end{equation}

The set $V^{(M)}$ contains all images ($I$) and associated data ($T$, e.g., text descriptions) in $\mathcal{K}$, and $v_j \in V$ is an entity containing relevant image-derived information and corresponds to image $I_j \in I$. The set $E^{(M)} = E^{(I)} \cup E^{(T)}$ contains the relations (simple edges or hyperedges) from image and text-derived information as in $V^{(M)}$. At retrieval time, we use our multimodal construction of $\mathcal{K}$ to retrieve the optimal subgraph $K^*_q$ with respect to a multimodal query $q=(I_q, T_q)$, where $I_q$ is an image and $T_q$ the query text. 

\subsection{Multimodal Knowledge Construction}
We can define $V^{(M)}$ by decomposing it into image ($I$) and text ($T$) components, as in Eqn. \eqref{eq:entity-def}. 
\begin{equation} \label{eq:entity-def}
    V^{(I)} = \rho_V(I), 
    V^{(T)} = \phi_V(\pi, T, p_{\text{ext}})
\end{equation}
% \begin{align} \label{eq:entity-def}
%     V^{(I)} &= \rho_V(I) \\
%     V^{(T)} &= \phi_V(\pi, T, p_{\text{ext}})
% \end{align}
%
The function $\rho$ is composed of image analysis modules, each returning different aspects of a particular $I_j$, e.g., embeddings, shape, color, detection, optical character recognition (OCR), etc. These qualities were chosen for their relevance in tobacco product analysis. For instance, color provides clues to flavors, a highly valuable data point. Legible words on product packaging can provide clues about the contents of packages, e.g., promotions/rewards. It allows for \textit{multi-pronged} knowledge and retrieval. The function $\phi_V$ takes care of extracting text descriptors for each $I_j$ given a LLM or LMM $\pi$, text $T$, and extraction prompt $p_{\text{ext}}$ to identify all entities within $T$. Thus, we have text information by which we can properly build a multimodal RAG framework.

The multimodal relations $E^{(M)}$ are similarly defined as the entity extraction in Equation \ref{eq:relation-def}. It relates entities to natural-language descriptions, e.g., text describing an image, and to other features such as color and shape.

\begin{equation} \label{eq:relation-def}
    E^{(I)} = \rho_E(I), 
    E_H^{(T)} = \phi_E(\pi, T, p_{\text{ext}})
\end{equation}
% \begin{align} \label{eq:relation-def}
%     E^{(I)} &= \rho_E(I) \\
%     E_H^{(T)} &= \phi_E(\pi, T, p_{\text{ext}})
% \end{align}

For $\rho_V$, we decompose it into separate extraction functions $\rho_V(I) = \cup_{j=1}\cup_{l=1}(\lambda_l(I_j))$ where $\lambda_l$ is a function that extracts one type of feature. Consequently, we define image-feature relations as $\rho_E(I) = \cup_{j=1}\cup_{l=1}(I_j, \lambda_l(I_j))$. In practice, $\phi$ uses the LMM $\pi$ to extract the text entities and relations in a typical fashion; querying $\pi$ for all $I_j$ would be prohibitively costly for $\phi$. Thus, our multimodal knowledge construction, broken down, is defined in Eqn. \eqref{eq:our-construction}.
\begin{equation} \label{eq:our-construction}
    \mathcal{K} = \Big(V^{(M)}, E_H^{(M)} \Big) = \Big(\rho_V(I) \cup \phi_V(\pi, T, p_{\text{ext}}), \rho_E(I) \cup \phi_E(\pi, T, p_{\text{ext}}) \Big) \\
\end{equation}
\subsection{Multimodal Knowledge Retrieval for Question Answering}

With our multi-pronged architecture, we can process multimodal queries according to different criteria, leveraging features that address public health needs (rather than merely text or image features). The image descriptors from $\phi$ are combined within the function $\mathcal{M}$. Our retrieval formulation is given in Eqn. \eqref{eq:our-retrieval-formulation}.
\begin{equation} \label{eq:our-retrieval-formulation}
    K_q^* = \mathcal{M}\{v \in V^{(M)} \cup \rho_V(I_q), e \in E^{(M)} \cup \rho_E(I_q) |{q,I_q}\} \cup K_{\text{chunk}}
\end{equation}
To grab entities from $\mathcal{K}$ at query time, we perform top-$k$ entity and relation matching. For images, we retrieve top-$k$ matches based on: (i) image embeddings; (ii) the image description; (iii) average color similarity; (iv) the shape description of the object; and (v) OCR contents, i.e., the similarity of text that is present in the image.
Finally, $\pi$'s final response to the user comes in the form $y^* = \pi(q | p_{\text{gen}}, K^*_q)$, where $p_{\text{gen}}$ is our response generation prompt.

\section{Experiments and Results}

\begin{figure}[tp]
    \centering
    % Define the box dimensions: e.g., width=\textwidth, height=5cm
    %\framebox[\textwidth]{\rule{0pt}{5cm}}
    \includegraphics[width=1.0\linewidth]{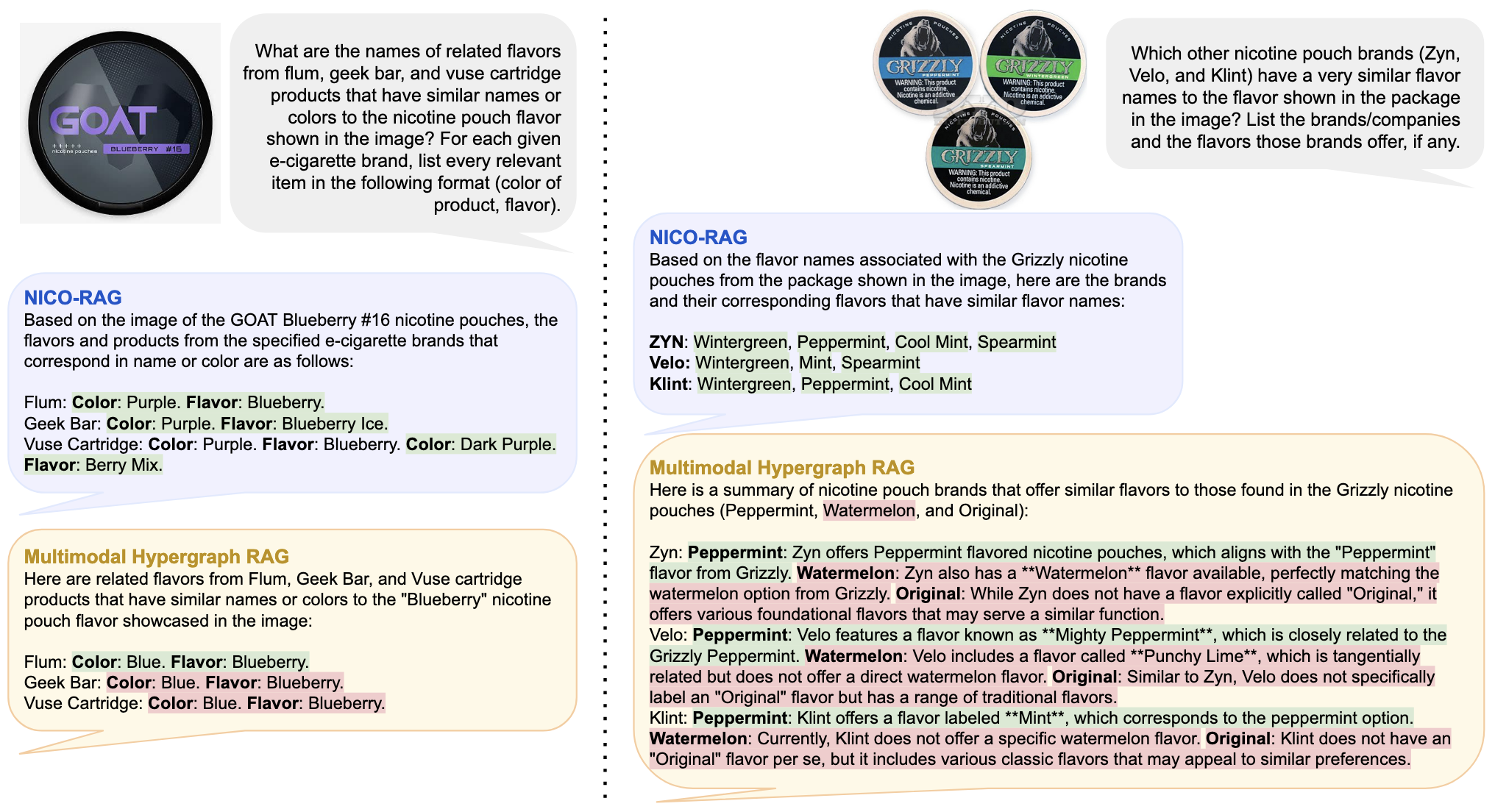}
    \caption{
        Responses from NICO-RAG and a multimodal Hypergraph RAG \cite{luo2025hypergraphrag} for two nicotine pouch products, where a complex search for inter- and intra-product information on image attributes is needed. 
        Best viewed in zoom and in color. Green highlights denote correct descriptions, while red denotes lower-quality or incorrect descriptions.
    }
    \label{fig:generation-vis}
\end{figure}

\subsection{Experimental Design} \label{subsec:experimental-design}
\textbf{Implementation.} We implement all our code using PyTorch in Python. As in prior work, we use GPT-4o-mini \cite{openai2024gpt4technicalreport} for $\pi$ due to its low cost; to process text within $\mathcal{K}$, we use \texttt{text-embedding-3-small}. For $\rho$’s extraction functions $\lambda_l$, we use CLIP ViT-14 \cite{radford2021learningtransferablevisualmodels} for image embeddings, DocTR \cite{doctr2021} for OCR, and Qwen3-VL (4B model) \cite{Qwen3-VL} for image descriptions. Top-$k$ retrieval is computed with cosine similarity. The image analysis modules were run on distributed computing servers equipped with GPUs ranging from Quadro RTX 8000 to A100. 

\noindent
\textbf{Test Data.} Our experiments focus on question-answering centered around nicotine pouch products. We sample 9 images from Zyn, 8 from Velo, and 5 from Klint, and 7 non-nicotine pouch products to simulate answering with existing knowledge. For unseen brands, we exclude all images collected under the Grizzly and Goat brands from $\mathcal{K}$, which produce smokeless tobacco but have now released pouch products. Furthermore, to mitigate the impact of noisy or mislabeled images, we manually removed any pouch product images for Grizzly and Goat that appeared under other brands, selecting the top 512 most similar samples. This amounts to approximately 3,700 images removed and 108 images in our test set. In total, $\pi$ will return 197 responses per experiment.

\noindent
\textbf{RAG Methods.} We use several RAG backbones $\pi$, adapted for the multimodal setting. 
\textit{\textbf{Naive generation}}: We simply query $\pi$ with $q$ and return its response as the answer. 
\textit{\textbf{Standard multimodal RAG}}: In this case, we retrieve the top-$k$ images, giving us corresponding image descriptions, which in practice are computed offline. 
\textit{\textbf{Naive multimodal HyperGraphRAG}}: An adaptation of HypergraphRAG \cite{luo2025hypergraphrag} where image descriptions are concatenated together and the entities and relations extracted, with images tied to corresponding chunks. We chose the hypergraph representation because it provides a strong natural-language representation of entities and relations in text-modality problems. 
\textit{\textbf{NICO-RAG}}: Our proposed NICO-RAG as discussed in Section \ref{sec:methodology}. We use all image analysis modules for multi-image feature retrieval rather than just embeddings. To connect image entities, we use the hypergraph structure from HyperGraphRAG for text chunks, and we tie all image descriptors to $\mathcal{K}$. 

\noindent
\textbf{Experiments.} In the first set of experiments, to measure response and golden answer word-level similarity, we use the F1 score. To measure semantic similarity, we follow \cite{luo2025hypergraphrag, es2025ragasautomatedevaluationretrieval} and use retrieval similarity (RS). To assess generation quality across multiple aspects, we score responses using an LLM as the judge \cite{luo2025hypergraphrag, que2024hellobench}. Evaluation ranges from inter-product questions (centered on pouches, with comparisons to non-pouch products) to intra-\textit{pouch}- brand questions. 
In the second set of experiments, we first conduct an ablation study on the different image analysis modules by omitting certain $\lambda_l$ using $k=4$ or $k=8$ top image descriptions for each $\lambda_l$. For our experiments, $\lambda_1$ gives color descriptions, $\lambda_2$ gives shape descriptions, $\lambda_3$ gives OCR predictions, and $\lambda_4$ gives image descriptions.

%\textcolor{red}{In the second set of experiments, we evaluate the impact of changing $k$ when retrieving the top-$k$ entities and edges, as well as limitations in retrieval length}. 
%In the second set of experiments, we evaluate the top-$k$ accuracy of image retrieval using different image analysis functions $\lambda_l$ within $\rho$. 
%In the fourth set of experiments, we ablate each component of our proposed method via an ablation study where different combinations of components are tested for retrieval. <--not necessary since exp1 and 2 take care of this

\subsection{Experimental Results and Discussion}

% Table 1: response generation evaluation via several scores
\begin{table*}[htb]\centering
\caption{
    Multimodal response evaluation of $\pi$’s responses given different multimodal RAG methods. Our results are in \textbf{bold}.
}
\label{tab:response-evaluation}
\begin{tabular}{lccc} 
\toprule
\textbf{RAG Method} & \textbf{F1} & \textbf{RS} & \textbf{GE} \\
\midrule
Naive Generation & 0.315 & 0.786 & 0.456 \\
Standard RAG & 0.295 & 0.807 & 0.473 \\
Hypergraph RAG & 0.283 & 0.789 & 0.467 \\
\midrule
\textbf{NICO-RAG} (Ours) & 0.273 & 0.800 & 0.466 \\
\bottomrule
\end{tabular}
\end{table*}

% Table 2: Image retrieval quality based on different combinations of the image analysis modules, thus affecting the confidence of retrieval
\begin{table*}[htb]\centering
\caption{
    Ablation study on NICO-RAG, where we showcase the contribution of each of the image analysis $\lambda$ functions.
}
\label{tab:image-retrieval-eval}
\begin{tabular}{lcccccc} 
\toprule
%\textbf{Included} & $k=1$ & $k=8$ & $k=16$ \\
\textbf{Included}  &    & $k=8$ &    &    & $k=4$ & \\
\textbf{Functions} & \textbf{F1} & \textbf{RS}    & \textbf{GE} & \textbf{F1} & \textbf{RS}     & \textbf{GE} \\
\midrule % 0.180 & 0.641 & 0.337 \\
$\{\lambda_1\}$                                   & 0.261 & 0.784 & 0.438 & 0.253 & 0.764 & 0.411 \\
$\{\lambda_1, \lambda_2\}$                        & 0.263 & 0.784 & 0.433 & 0.250 & 0.769 & 0.413 \\
$\{\lambda_1, \lambda_2, \lambda_3 \}$            & 0.269 & 0.784 & 0.446 & 0.257 & 0.782 & 0.432 \\
$\{\lambda_1, \lambda_2, \lambda_3, \lambda_4 \}$ & 0.273 & 0.800 & 0.466 & 0.261 & 0.792 & 0.448 \\
\bottomrule
\end{tabular}
\end{table*}

Table \ref{tab:response-evaluation} shows the results of the generation evaluation across different RAG methods. The multimodal HypergraphRAG, with three calls to $\pi$ (including two that use $I_q$), achieves similar performance to our NICO-RAG, which removes one call to $\pi$ entirely in favor of $\lambda_l$, and only one call involves $I_q$. This shows that, in large-scale scenarios, we need not rely on the expressive power of $\pi$ to obtain accurate image-level information. Meanwhile, Table \ref{tab:image-retrieval-eval} shows us the effects of providing the top-$k$ sample descriptions from image retrieval. $\lambda_1$ returns color descriptors, $\lambda_2$ returns shape descriptors, $\lambda_3$ returns OCR terms, and $\lambda_4$ returns an image description of the corresponding $I_q$. 

\section{Conclusion}
In this work, we introduced the NICO dataset, a collection of nicotine and tobacco product samples of over 200,000 samples. To the best of our knowledge, it is the largest and most diverse dataset of its kind. We also presented NICO-RAG, a RAG framework that leverages diverse image descriptors to construct a multimodal hypergraph-based knowledge base while reducing dependency on heavy LMMs. From our experiments with NICO-RAG and the adaptation of the state-of-the-art text RAG method, we observe comparable performance without adding an additional LMM call per query. At the same time, NICO-RAG can also return factual responses even with the removal of image-based entity extraction from $\pi$.
Despite its performance, NICO-RAG has avenues of improvement. The heavier $\lambda_l$ functions, such as OCR and descriptions via Qwen3-VL, still require relatively expensive non-consumer grade hardware to run at the scale of NICO. 
We argue, however, that large models such as $\pi$ cannot be run locally and require monetary resources, unlike our proposed $\lambda_l$ functions. 
Furthermore, constructing $\mathcal{K}$ from image and text descriptions can be prohibitively time-consuming and costly if we rely solely on $\pi$ to perform these tasks, necessitating the manual insertion of relations and entities, as in our method (i.e., the image modules).
Despite our efforts, the NICO dataset contains images that are not relevant to tobacco or nicotine product research, i.e., noisy samples. We argue that this is mitigated by our multiple retrieval criteria per image, in which irrelevant images are weighted less than relevant samples. This is especially the case with OCR and image descriptions. 
We aim to foster further work to harmonize large-scale datasets with the expressive power of RAG frameworks, with the goal of resolving public health crises and yielding results more quickly.

\bibliographystyle{splncs04}
\bibliography{main}
\end{document}